# Siamese Neural Networks for One-shot detection of Railway Track Switches

**Dattaraj J Rao, Shruti Mittal, S. Ritika,**
General Electric

## ABSTRACT

Deep Learning methods have been extensively used to analyze video data to extract valuable information by classifying image frames and detecting objects. We describe a unique approach for using video feed from a moving Locomotive to continuously monitor the Railway Track and detect significant assets like Switches on the Track. The technique used here is called Siamese Networks – which uses 2 identical networks to learn the similarity between of 2 images. Here we will use a Siamese network to continuously compare Track images and detect any significant difference in the Track. Switch will be one of those images that will be different and we will find a mapping that clearly distinguishes the Switch from other possible Track anomalies. The same method will then be extended to detect any abnormalities on the Railway Track.

Railway Transportation is unique in the sense that is has wheeled vehicles – Trains pulled by Locomotives - running on guided Rails at very high speeds nearing 200 mph. Multiple Tracks on the Rail network are connected to each other using an equipment called Switch or a Turnout. Switch is either operated manually or automatically through command from a Control center and it governs the movement of Trains on different Tracks of the network. Accurate location of these Switches is very important for the railroad and getting a true picture of their state in field is important. Modern trains use high definition video cameras facing the Track that continuously record video from track. Using a Siamese network and comparing to benchmark images – we describe a method to monitor the Track and highlight anomalies.

## INTRODUCTION

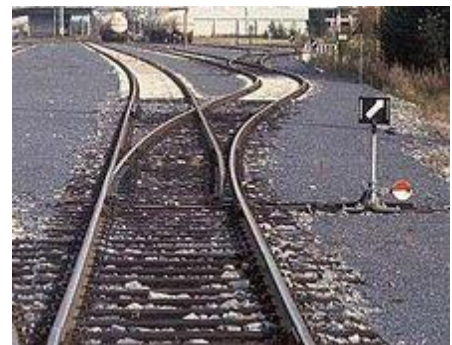

A railroad switch or turnout is a mechanical installation enabling railway trains to be guided from one track to another, such as at a railway junction or where a spur or siding branches off. The switch consists of the pair of linked tapering rails, known as points (switch rails or point blades), lying between the diverging outer rails (the stock rails). These points can be moved laterally into one of two positions to direct a train coming from the point blades toward the straight path or the diverging path. A train moving from the narrow end toward the point blades (i.e. it will be directed to one of the two paths depending on the position of the points) is said to be executing a facing-point movement. Unless the switch is locked, a train coming from either of the converging directs will pass through the points onto the narrow end, regardless of the position of the points, as the vehicle's wheels will force the points to move. Passage through a switch in this direction is known as a trailing-point movement.

Siamese networks are Convolutional Neural Networks that map an input space (image array) to a vector space that tries to capture the internal patterns in an image. Comparing these vector spaces generated from 2 images using the exact same network (same structure and exact same weights) gives us a measure of the similarity between the 2 images. This concept is used to develop One-Shot Detector models that can learn on limited examples. Instead of regular Convolutional Neural Networks – these One-shot Detector networks can be trained with limited training data. The end objective is not to exactly classify an image but to extract a vector representation that gives good idea of the similarity of the image with some established benchmark.

## Applying to Track detection problem

Modern Locomotives have a high-definition Video camera that continuously records Track Video. Below is a sample of this video collected from GE Transportation's LocoVISION System. The system has a specially Engineered HD camera that accounts for Train vibrations and records a steady image.

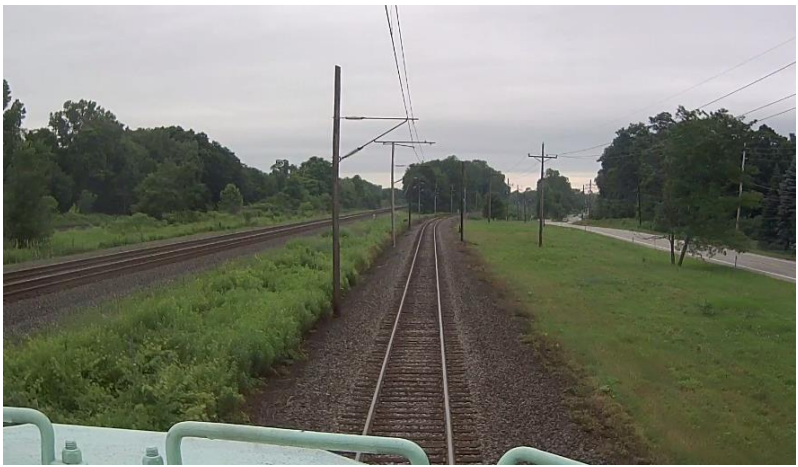

The videos are typically collected at 30 frames per second. So we would have 30 images every second which we can analyze. During real-world analysis, however we may decide to use a frequency of analysis lower than this – in range of 5 or 10 frames per second. This is because even at speeds of 200 mph we will be able to cover most of the Track at 10 frames per second. Also this will lower our processing power requirement in case this analysis happens at the Edge – that is on an onboard computer on a Locomotive.

The other option is to upload these videos periodically to Cloud using WiFi – when the Train approaches a region with good WiFi connectivity – example Yards or Stations. During this time, a bulk upload is done and once data is ingested in Cloud an analysis is done. The Deep Learning models are deployed the Cloud and can be run on a cluster of CPUs or GPUs. The parallelization helps us improve performance and get back insights in time.

# Data Collection for Training

We sample the image to get only the Track as region of interest. Using the video, we take multiple samples of the Track as shown in the figure below. These samples serve as training images for the network. We collect around 500 samples from the video from different regions where the Trains run. We will collect images with the Track selected as region of interest and label these as "Positive". We collect images from different lighting conditions during different times of the day. Now we use these training images to train the network. The images shown to left show a Track and will be considered as Positives for our case study. These contain the Track.

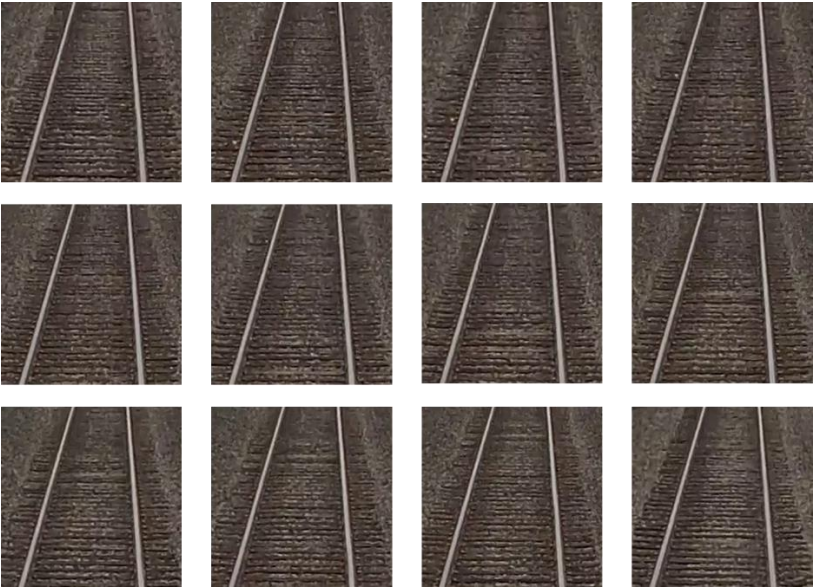

Now we collect images from a different section of the video and consider them as Negatives for our Training – since these do not contain the Track.

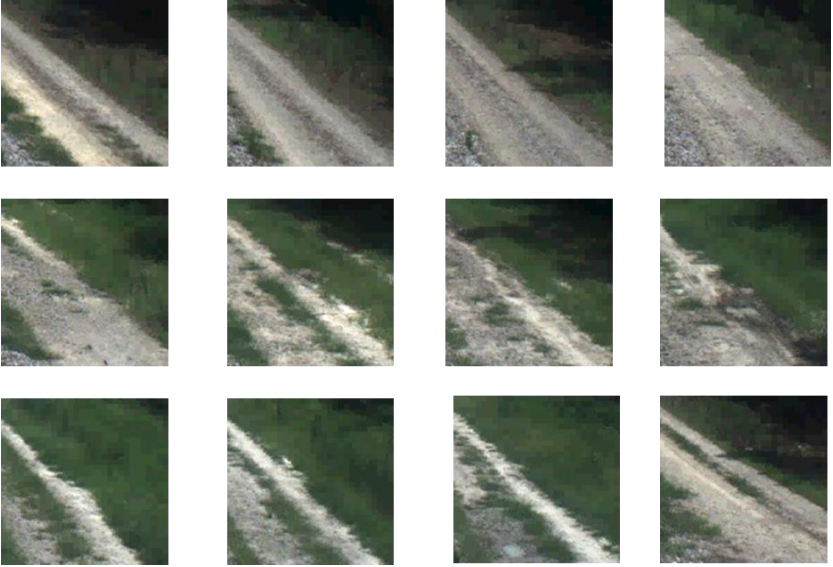

# Building the Siamese Network

The Siamese network uses 2 Convolutional networks with same structure and weights. We will explain a method used for our specific problem here – there may be different/better options for implementing Siamese networks. By no means do we claim this to be the best.

Below figure shows the structure of Siamese network we use.

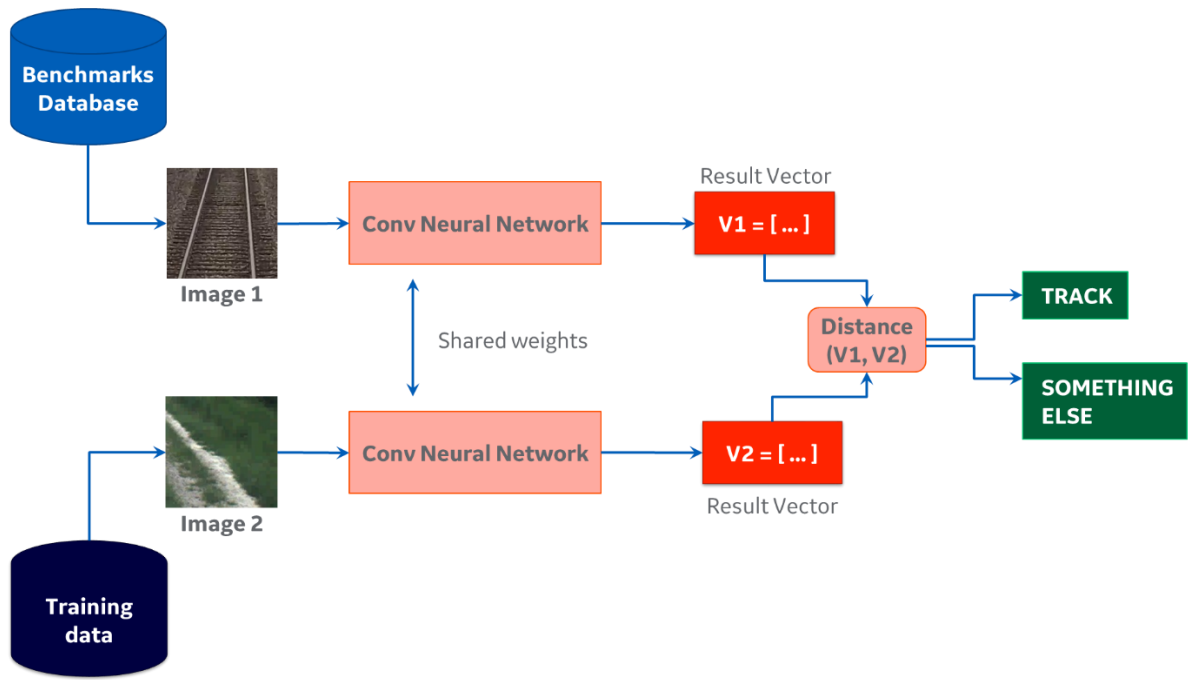

We select images with the best railway Track as our benchmark. We will compare Track images with the benchmarks and predict how different they are – and highlight ones with significant difference as anomalies. We are not predicting why they are different but only that they are significantly different from the benchmarks.

The images we collected as positive and negatives are used for training the network. We take each benchmark, combine alternatively with positive and negative samples and send the tuple as a training example. We add a label for the training example as 0 or 1 – indicating Euclidean distance with benchmark images. For our case – 0 indicates high similarity and 1 indicates maximum distance and thus very different image. Here we give 0 as label for our positive images and for the negatives we give label as 1.

We train the model to an accuracy of 95%. Now as we feed new image to the network and compare to benchmark image – it will output a value between 0 and 1 depending on presence of a Track. We find that this network can detect occurrence of anomalies on the Track – anything that does not look like a Track. This includes any portion of the Track that looks different like road crossing, switches, loose ballast, etc.

Below are examples of some images that were detected by the network as anomalies.

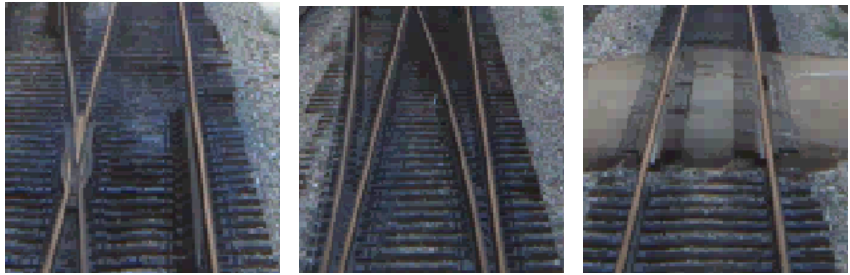

Below figure shows how the system works once deployed. We will analyze Track images and select the Track region and feed the Track images to the network. The network will analyze these and highlight significant differences – these must then be analyzed by humans to find out if they are real anomalies.

If the system must be deployed to a new region or terrain – these are significant differences in the way the Tracks and terrains look. For example, the railway Track in India is significantly different from the same in USA. In such cases – we collect few hundred images from the new terrain video and re-Train the model to start looking for anomalies.

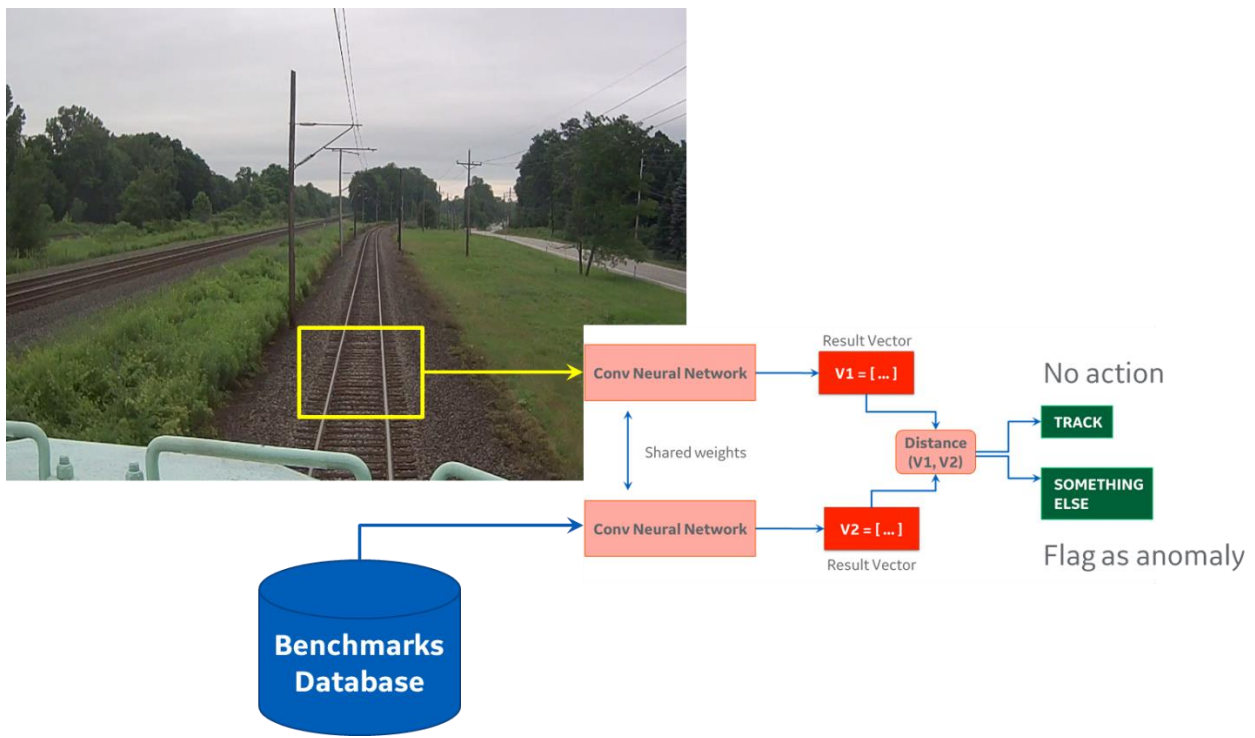

# CONCLUSION

We saw the use of a Siamese Neural Network for Training on Railway Track images and learning to predict images that "look" different from a regular Railway Track image. This network can now start analyzing live data from Train video camera being collected at 30 frames per second – and alert on presence of anomalies like road-crossings, switches, etc.

Compared to regular Deep Learning networks – this is a very lightweight network. Also when we move to a new region the Tracks tend to be very different. Here we can collect a few Training examples of the new Track and re-Train the model of the region. Then it will start detecting anomalies for the new region.

Following are the key advantages provided by this approach:

- **Performance**: Low footprint model – can run on Edge or Cloud
- **Scalable**: Detect new types of anomalies – not only what its trained for
- **Extendable**: Can be easily re-trained on few images from a new Track/Region